\pdfoutput=1
\documentclass[runningheads]{llncs}
\usepackage{graphicx}
\begin{document}
\title{The Role of Bio-Inspired Modularity in General Learning\thanks{Supported by Machine Perception and Cognitive Robotics Labs}}
%
%
\author{Rachel A. StClair \inst{1} \and
William Edward Hahn \inst{1} \and
Elan Barenholtz \inst{1}}
\authorrunning{R. StClair et al.}
%
\institute{Center for Complex Systems and Brain Sciences and Center for Future Mind, Florida Atlantic University, Florida, USA\\ 
\email{\{rstclair2012\}@fau.edu}}
\maketitle              
\begin{abstract}
One goal of general intelligence is to learn novel information without overwriting prior learning. The utility of learning without forgetting (CF) is twofold: first, the system can return to previously learned tasks after learning something new. In addition, bootstrapping previous knowledge may allow for faster learning of a novel task. Previous approaches to CF and bootstrapping are primarily based on modifying learning in the form of changing weights to tune the model to the current task, overwriting previously tuned weights from previous tasks. However, another critical factor that has been largely overlooked is the initial network topology, or architecture. Here, we argue that the topology of biological brains likely evolved certain features that are designed to achieve this kind of informational conservation. In particular, we consider that the highly conserved property of modularity may offer a solution to weight-update learning methods that adheres to the learning without catastrophic forgetting and bootstrapping constraints. Final considerations are then made on how to combine these two learning objectives in a dynamical, general learning system.

\end{abstract}
\section{Introduction}

\subsection{Knowledge Preservation}
A basic goal of most approaches to artificial general intelligence is the ability to learn multiple tasks without overwriting what has previously been learned (i.e. catastrophic forgetting, or CF). The potential utility of learning without forgetting, or 'knowledge preservation', is twofold: first, the system can return to previously learned tasks after learning something new. In addition, bootstrapping previous knowledge may allow for faster learning of a novel task through transfer learning. Both of these potential benefits of preserving learned information may serve to conserve resources. Allowing the same computational hardware to be ‘multiplexed’ for another task, rather than requiring allocation of distinct resources which reduces the required energy resources involved in the learning process. In the former, multiplexed information is stored in such a way that the system can switch between already learned tasks while retaining performance. In the latter case, information is accessed for learning new tasks in a contextually relevant manner. Multiplexed information is thus stored and retrieved in a way that facilitates the conservation and use of prior learning, broadening the ability of a learning system to multiple task domains.

\subsection{Previous Approaches}
A significant body of recent work has explored approaches to CF and bootstrapping in relation to multi-layer neural networks or deep learning (DL) \cite{xie2020artificial}. Learning in these systems consists of updating the weights between nodes according to their respective partial derivatives calculated from a loss function (i.e. backpropagation). With the recent implementation of auto-differentiation and parallel computing devices, progress in backpropagation-based DL has rapidly emerged as the dominant approach in AI \cite{schmidhuber2015deep}. Previous approaches to overcoming CF in DL models have involved strategies with regard to the learning process itself. Most commonly in the case of bootstrapping, ensemble networks are re-trained on resampled datasets with little improvement \cite{nixon2020bootstrapped}. Transfer learning is also used as a bootstrapping method in which previously-learned features are imported as the starting weights for later update \cite{weiss2016survey}. CF approaches are more diverse, mostly involving modification to the weight update methods. Notably, Elastic Weight Consolidation has shown promising results in overcoming CF in sequences of Atari tasks by slowing down updates on previously encoded weights \cite{kirkpatrick2017overcoming,aich2021elastic}. However, these approaches have not been widely adopted, likely because the techniques do not demonstrate knowledge preservation on a level necessary for general intelligence. We suspect this is in part due to the fundamental nature of backpropagation in DL models to rewrite learning; where weights are updated most heavily towards current learning at the expense of weights learned in more distant tasks.

\subsection{Network Topology}
Another potential approach that has thus far been largely neglected  is to consider the topology of the learning system — that is, its ‘architecture’. While there is a significant body of work on architecture optimization for other metrics of learning \cite{lukovsevivcius2012reservoir,elsken2019neural,goudarzi2012emergent,vanschoren2018meta,lee2014evolution,williams2017evolution,scholkopf2021toward}, to date there has been little work considering architectural choices in relation to CF and/or bootstrapping. Fortunately, hundreds of million years of evolution has generated highly conserved, vertebrate and mammalian brain architectures that have been optimized with regard to intelligent functions including, it is reasonable to assume,  the preservation of learning without CF.

One potentially promising property of biological  nervous system architectures, which may serve to overcome the pitfalls of CF, is the presence of modular architectures such as the cortex and subcortical structures. Modularity is a highly conserved property that is present across many organisms with general learning abilities. Complex neural architectures are thought to have emerged in the Cambrian explosion, 540 million years ago, with records of both camera eyes and the emergence of animals with complex behaviors \cite{trestman2013cambrian}. In contemporary species, the four basic brain structures (Telencephalon, Cerebellum, Diencephalon, Mesencephalon) have been preserved across nearly all mammalian species and most avian, reptile, shark brains, and some fish brains \cite{avianB,Yopak12946,Kotrschal2020,larsell1926cerebellum,b9}. Insects have their own highly preserved modular nervous-system  structures as well \cite{ITO2014755}.

In humans, modularity is highly pervasive. For example, the cortex processes incoming sensory information while the basal ganglia controls motor movements in behavioral learning \cite{ring2002neuropsychiatry,sutherland1988contributions}. These structures are specialized not only with regard to the types of information they  processed (e.g. photon patterns for V1 stream vs. layers 5/6 cortical output from cortex), but also with regard to their neuron types as well as their morphological structures. Communication between the two structures occurs in specialized. limited routes, cortico-basal ganglia networks, which serve to integrate information between the two regions \cite{lalo2008patterns}. The result of this particular organization is a cortex that is optimized for sensory information, while the basal ganglia is optimized to interpret cortical information for motor movement implementation \cite{coward2005system}.

Here, we propose that such modular architectures may also serve to preserve learning during the update process by separating different types of computation (specialization) and storing the learning across un-updatable boundaries (segregation) which communicate through specific upstream pathways (integration). In artificial systems, this modularity  may be modeled in the form of segregation of weight updates based on local update rules within a module. Previous work on artificial systems has looked at the effects of modularity for generalization, but using different learning metrics \cite{goyal2019recurrent,goyal2020inductive,werle2020investigating}. Promising results on the effects of modularity for generalizing to out-of-distribution data has been shown in the Badger architecture, which relies on nested self-sufficient computational units which communicate for a shared objective \cite{rosa2019badger}. Badger is an agent based model in the sense that each agent, or small self-sufficient network, updates in an internal loop, while the outer loop optimizes the collection of homogeneous agents. However, badger does not incorporate within-agent modularity which we believe is essential to solving the knowledge preservation problem. 

While we have attempted to make the specific case for modularity as an example of a potentially important design principle in natural systems, we propose that knowledge preservation should serve as a general guiding framework in architecture design that may give rise to dynamical learning systems capable of some level of generality. The key idea is modeled from the human brain where it is likely that network topology is perhaps at least, if not more important than later learning algorithms in achieving this goal. This allows us to avoid ending up with a system that is doing the bulk of the learning process after it has been configured. In a system that prioritizes initial connectivity, this manifests as selecting appropriate modularity over a variety of different domain tasks. Rewriting existing learning prevents sharing of computational resources for novel tasks. This is often the problem in designing AGI from theoretical systems that don’t take resource constraints into account, such as universal Turing machines and AIXI \cite{turing2009computing,hutter2003gentle}.

\section{Possible Implementation Approaches}
Many methods could be conceived for discovering the inductive biases that modularity has to offer learning system architectures. Crucially, there is a tradeoff between initial connectivity and later weight modification. A strong dynamical learning system would utilize initial topology search as a prior learning experience to later be exploited during the weight modification learning phase.

Consider regions separated by un-updatable boundaries that delineate the extent of backpropagation. These modules could optimize without overwriting each other, while still being able to share resources when relevant (i.e. bootstrapping). This approach also limits rewriting prior learning for the sake of the current learning, where modules are activated at different times depending on their relevance to the current tasks. Furthermore, modules dedicated to computational tasks, instead of behavioral tasks (e.g. memory, sensory, self-activation, etc.) may serve as a more biologically inspired approach, since this is largely what we observe between cortical and subcortical structures.

Evolutionary searches are particularly useful in discovering unique solutions to optimization problems, which can be extended here. Although computationally extravagant, one promising candidate would be an evolutionary approach that couples network topology and embodiment search to different learning environments that also evolve over time. Derivatives of this approach would be useful, especially those focusing on modularity and learning. The advantage to this method is that appropriate submodules will likely be paired with the particular sensory receptors given to the agent and the types of learning that needs to be delegated throughout the system in ways that may not be currently known.

Other approaches include intelligently designed architectures where architecture is random, learned, selected by an operator, and most commonly in combination. Random networks, such as \cite{gershenson2004introduction}, have seen interesting results and may offer some potential insights into the core properties of learning, however they will likely  have to be harnessed in a constructive way to get desired behavior, as in \cite{lukovsevivcius2012reservoir}. Similar to random learning, methods exist for crystallizing order from disorder in unorganized networks as in the case of \cite{gutowitz1991cellular,hebb1949organisation}, leading to module-like structures. Still, there are other conceivable approaches involving learning functions that select sequences of computational architecture pathways and more. More extensive studies could iteratively search through all possible configurations of N-node networks to discover what modularity benefits particular learning environments. Even hand-designed simple models that start to test the questions of module design and learning without forgetting would be useful in answering the questions presented here.

\section{Benchmarking Generality}
In order to assess a model's knowledge preservation, it must be able to demonstrate that it can learn a sequence of tasks in a variety of domains and return to those tasks with improvements in computational costs and/or learning time. Below we describe initial tests to measure generality in terms of learning without forgetting and bootstrapping. However, as generality scales, more advanced testing is likely necessary.

\subsubsection{Knowledge Preservation Metrics}
Learning without catastrophic forgetting is typically  measured by performance on a sequence of tasks. Here, we propose it is imperative to test on sequences in different domains (e.g. vision classification and time-series forecasting) to demonstrate generality over narrow-learning.

\subsubsection{Boostrapping A-Priori Knowledge Metrics}
Measuring bootstrapping a-priori knowledge is slightly more challenging due to the effects seen in transfer learning. The goal here is to see an increased speed in learning task B after learning task A. Again, in trying to capture generality, it is useful to compare tasks from different domains so the distribution of features in the data is different enough that it's possible to measure using contextual knowledge instead of the re-use of features. This later point is what occurs in transfer learning. As in the case of the popular technique of using pre-trained ResNet on Imagenet and vision based tasks, the transferred weights capture many features which would have to have been learned in the secondary network either way \cite{weiss2016survey}. Here, we also want to be able to use learned features to aid in learning a new task, but we want to ensure that these are general features that can be maximally shared across all tasks the system may be expected to learn. Teasing out this difference is aided in investigating how data flows through different parts of the system’s topology. For example, the Fusiform Face Area is shown to activate when distinguishing between faces and also in bird experts for distinguishing between birds \cite{tong2008fusiform}. This multi-use activation of the same subnetwork is one example of bootstrap learning. We might expect similar submodules to be redeployed across different kinds of learning domains.

\subsubsection{Challenges}
We recognize that adding modularity alone will probably not multiplex information in a way that provides human-level general intelligence. The works of Coward, Buszaki, and Turing suggest algorithms employed by the human brain may give rise to this generality based on eliciting sequences of computation across different modules  \cite{coward2005system,gyorgy2019brain,turing1969intelligent}. Furthermore, special message passing algorithms between the regions would benefit from that of publish and subscribe models; that regions in the system can communicate with each other when necessary.

\section{Conclusion}
We have highlighted topology as an important topic for future research, as well as the need for algorithms that allow for learning without rewriting previous learning. We suspect modularity in biological intelligent systems plays a crucial role in achieving this need while conserving resources. This paper also contributes to discussions on AGI agent architectures by suggesting some useful considerations on how to test for generality in regards to CF and bootstrapping. Works from \cite{wang2019paired,rosa2019badger,koutnik2013evolving,stanley2019designing,aich2021elastic} have attempted to tackle these issues separately, while others \cite{silver2021reward,wang2021modular} have come from the angle of single-use systems, all of which overwrite prior learning to a unsatisfactory degree. Still, some works have come from the position of down-scaling theoretical models without resource constraints \cite{goertzel2007artificial,hutter2003gentle,turing2009computing}. While these approaches are useful, we find the proposed approach is more appropriate for the pursuit of general intelligence. We believe this paper provides a framework for investigating computationally tractable general learning systems. Promising directions involve evolutionary and intelligent topology design that utilizes modularity to minimize rewriting prior learning. We hope this paper serves as a tool for those interested in benchmarking early progress in generality. There is room for many future works to provide insight into the approach presented here as well as how to measure generality in terms of bootstrapping and learning new without forgetting old. We have tried to clarify how modularity may offer a method of learning without rewriting and the role it plays in topology design. Much work needs to be done to improve that understanding and find ways to appropriately incorporate modularity in AGI frameworks.

\bibliographystyle{splncs04}
%
\bibliography{paper}

\begin{thebibliography}{10}
\providecommand{\url}[1]{\texttt{#1}}
\providecommand{\urlprefix}{URL }
\providecommand{\doi}[1]{https://doi.org/#1}

\bibitem{aich2021elastic}
Aich, A.: Elastic weight consolidation (ewc): Nuts and bolts. arXiv preprint
  arXiv:2105.04093  (2021)

\bibitem{b9}
Bruce, L.L.: Evolution of the Brain in Reptiles, pp. 1295--1301. Springer
  Berlin Heidelberg, Berlin, Heidelberg (2009)

\bibitem{coward2005system}
Coward, L.A.: A system architecture approach to the brain: From neurons to
  consciousness. Nova Publishers (2005)

\bibitem{elsken2019neural}
Elsken, T., Metzen, J.H., Hutter, F.: Neural architecture search: A survey. The
  Journal of Machine Learning Research  \textbf{20}(1),  1997--2017 (2019)

\bibitem{gershenson2004introduction}
Gershenson, C.: Introduction to random boolean networks. arXiv preprint
  nlin/0408006  (2004)

\bibitem{goertzel2007artificial}
Goertzel, B., Pennachin, C.: Artificial general intelligence, vol.~2. Springer
  (2007)

\bibitem{goudarzi2012emergent}
Goudarzi, A., Teuscher, C., Gulbahce, N., Rohlf, T.: Emergent criticality
  through adaptive information processing in boolean networks. Physical review
  letters  \textbf{108}(12),  128702 (2012)

\bibitem{goyal2020inductive}
Goyal, A., Bengio, Y.: Inductive biases for deep learning of higher-level
  cognition. arXiv preprint arXiv:2011.15091  (2020)

\bibitem{goyal2019recurrent}
Goyal, A., Lamb, A., Hoffmann, J., Sodhani, S., Levine, S., Bengio, Y.,
  Sch{\"o}lkopf, B.: Recurrent independent mechanisms. arXiv preprint
  arXiv:1909.10893  (2019)

\bibitem{gutowitz1991cellular}
Gutowitz, H.: Cellular automata: Theory and experiment. MIT press (1991)

\bibitem{gyorgy2019brain}
Gy{\"o}rgy~Buzs{\'a}ki, M.: The brain from inside out. Oxford University Press
  (2019)

\bibitem{hebb1949organisation}
Hebb, D.O.: The organisation of behaviour: a neuropsychological theory. Science
  Editions New York (1949)

\bibitem{hutter2003gentle}
Hutter, M.: A gentle introduction to the universal algorithmic agent aixi
  (2003)

\bibitem{ITO2014755}
Ito, K., Shinomiya, K., Ito, M., Armstrong, J.D., Boyan, G., Hartenstein, V.,
  Harzsch, S., Heisenberg, M., Homberg, U., Jenett, A., Keshishian, H.,
  Restifo, L.L., Rössler, W., Simpson, J.H., Strausfeld, N.J., Strauss, R.,
  Vosshall, L.B.: A systematic nomenclature for the insect brain. Neuron
  \textbf{81}(4),  755--765 (2014).
  \doi{https://doi.org/10.1016/j.neuron.2013.12.017},
  \url{https://www.sciencedirect.com/science/article/pii/S0896627313011781}

\bibitem{kirkpatrick2017overcoming}
Kirkpatrick, J., Pascanu, R., Rabinowitz, N., Veness, J., Desjardins, G., Rusu,
  A.A., Milan, K., Quan, J., Ramalho, T., Grabska-Barwinska, A., et~al.:
  Overcoming catastrophic forgetting in neural networks. Proceedings of the
  national academy of sciences  \textbf{114}(13),  3521--3526 (2017)

\bibitem{Kotrschal2020}
Kotrschal, A., Kotrschal, K.: Fish Brains: Anatomy, Functionality, and
  Evolutionary Relationships, pp. 129--148. Springer International Publishing,
  Cham (2020)

\bibitem{koutnik2013evolving}
Koutn{\'\i}k, J., Cuccu, G., Schmidhuber, J., Gomez, F.: Evolving large-scale
  neural networks for vision-based reinforcement learning. In: Proceedings of
  the 15th annual conference on Genetic and evolutionary computation. pp.
  1061--1068 (2013)

\bibitem{lalo2008patterns}
Lalo, E., Thobois, S., Sharott, A., Polo, G., Mertens, P., Pogosyan, A., Brown,
  P.: Patterns of bidirectional communication between cortex and basal ganglia
  during movement in patients with parkinson disease. Journal of Neuroscience
  \textbf{28}(12),  3008--3016 (2008)

\bibitem{larsell1926cerebellum}
Larsell, O.: The cerebellum of reptiles: lizards and snake. Journal of
  Comparative Neurology  \textbf{41}(1),  59--94 (1926)

\bibitem{lee2014evolution}
Lee, D.S.: Evolution of regulatory networks towards adaptability and stability
  in a changing environment. Physical Review E  \textbf{90}(5),  052822 (2014)

\bibitem{lukovsevivcius2012reservoir}
Luko{\v{s}}evi{\v{c}}ius, M., Jaeger, H., Schrauwen, B.: Reservoir computing
  trends. KI-K{\"u}nstliche Intelligenz  \textbf{26}(4),  365--371 (2012)

\bibitem{werle2020investigating}
Werle van~der Merwe, A.: Investigating the evolution of modularity in neural
  networks. Ph.D. thesis, Stellenbosch: Stellenbosch University (2020)

\bibitem{nixon2020bootstrapped}
Nixon, J., Lakshminarayanan, B., Tran, D.: Why are bootstrapped deep ensembles
  not better? In: ''I Can't Believe It's Not Better!''NeurIPS 2020 workshop
  (2020)

\bibitem{avianB}
Nomura, T., Izawa, E.I.: Avian brains: Insights from development, behaviors and
  evolution. Development, Growth \& Differentiation  \textbf{59}(4),  244--257
  (2017). \doi{https://doi.org/10.1111/dgd.12362},
  \url{https://onlinelibrary.wiley.com/doi/abs/10.1111/dgd.12362}

\bibitem{ring2002neuropsychiatry}
Ring, H., Serra-Mestres, J.: Neuropsychiatry of the basal ganglia. Journal of
  Neurology, Neurosurgery \& Psychiatry  \textbf{72}(1),  12--21 (2002)

\bibitem{rosa2019badger}
Rosa, M., Afanasjeva, O., Andersson, S., Davidson, J., Guttenberg, N.,
  Hlubu{\v{c}}ek, P., Poliak, M., V{\'\i}tku, J., Feyereisl, J.: Badger:
  Learning to (learn [learning algorithms] through multi-agent communication).
  arXiv preprint arXiv:1912.01513  (2019)

\bibitem{schmidhuber2015deep}
Schmidhuber, J.: Deep learning in neural networks: An overview. Neural networks
   \textbf{61},  85--117 (2015)

\bibitem{scholkopf2021toward}
Sch{\"o}lkopf, B., Locatello, F., Bauer, S., Ke, N.R., Kalchbrenner, N., Goyal,
  A., Bengio, Y.: Toward causal representation learning. Proceedings of the
  IEEE  \textbf{109}(5),  612--634 (2021)

\bibitem{silver2021reward}
Silver, D., Singh, S., Precup, D., Sutton, R.S.: Reward is enough. Artificial
  Intelligence p. 103535 (2021)

\bibitem{stanley2019designing}
Stanley, K.O., Clune, J., Lehman, J., Miikkulainen, R.: Designing neural
  networks through neuroevolution. Nature Machine Intelligence  \textbf{1}(1),
  24--35 (2019)

\bibitem{sutherland1988contributions}
Sutherland, R., Whishaw, I., Kolb, B.: Contributions of cingulate cortex to two
  forms of spatial learning and memory. Journal of Neuroscience  \textbf{8}(6),
   1863--1872 (1988)

\bibitem{tong2008fusiform}
Tong, M.H., Joyce, C.A., Cottrell, G.W.: Why is the fusiform face area
  recruited for novel categories of expertise? a neurocomputational
  investigation. Brain research  \textbf{1202},  14--24 (2008)

\bibitem{trestman2013cambrian}
Trestman, M.: The cambrian explosion and the origins of embodied cognition.
  Biological Theory  \textbf{8}(1),  80--92 (2013)

\bibitem{turing1969intelligent}
Turing, A.: Intelligent machinery. 1948. The Essential Turing p.~395 (1969)

\bibitem{turing2009computing}
Turing, A.M.: Computing machinery and intelligence. In: Parsing the turing
  test, pp. 23--65. Springer (2009)

\bibitem{vanschoren2018meta}
Vanschoren, J.: Meta-learning: A survey. arXiv preprint arXiv:1810.03548
  (2018)

\bibitem{wang2021modular}
Wang, J., Elfwing, S., Uchibe, E.: Modular deep reinforcement learning from
  reward and punishment for robot navigation. Neural Networks  \textbf{135},
  115--126 (2021)

\bibitem{wang2019paired}
Wang, R., Lehman, J., Clune, J., Stanley, K.O.: Paired open-ended trailblazer
  (poet): Endlessly generating increasingly complex and diverse learning
  environments and their solutions. arXiv preprint arXiv:1901.01753  (2019)

\bibitem{weiss2016survey}
Weiss, K., Khoshgoftaar, T.M., Wang, D.: A survey of transfer learning. Journal
  of Big data  \textbf{3}(1),  1--40 (2016)

\bibitem{williams2017evolution}
Williams, S., Yaeger, L.: Evolution of neural dynamics in an ecological model.
  Geosciences  \textbf{7}(3), ~49 (2017)

\bibitem{xie2020artificial}
Xie, Z., He, F., Fu, S., Sato, I., Tao, D., Sugiyama, M.: Artificial neural
  variability for deep learning: On overfitting, noise memorization, and
  catastrophic forgetting. Neural Computation pp. 1--30 (2020)

\bibitem{Yopak12946}
Yopak, K.E., Lisney, T.J., Darlington, R.B., Collin, S.P., Montgomery, J.C.,
  Finlay, B.L.: A conserved pattern of brain scaling from sharks to primates.
  Proceedings of the National Academy of Sciences  \textbf{107}(29),
  12946--12951 (2010). \doi{10.1073/pnas.1002195107},
  \url{https://www.pnas.org/content/107/29/12946}

\end{thebibliography}
\end{document}